\documentclass[11pt,a4paper]{article}
\usepackage[hyperref]{eacl2021}
\usepackage{times}
\usepackage{latexsym}
\usepackage{times}
\usepackage{latexsym}
\usepackage{times}
\usepackage{soul}
\usepackage{url}
\usepackage{inputenc}
\usepackage{caption}
\usepackage{graphicx}
\usepackage{amsmath}
\usepackage{amsthm}
\usepackage{booktabs}
\usepackage{subfigure}
\usepackage{float}
\usepackage{amssymb}
\usepackage{times}
\usepackage{latexsym}
\usepackage{multirow}
\usepackage{booktabs}
\usepackage{tabularx}
\usepackage{graphics}
 \usepackage{graphicx}
 \usepackage{amssymb}
\usepackage[ruled,vlined]{algorithm2e}

\usepackage{microtype}

\aclfinalcopy 

\newcommand*\samethanks[1][\value{footnote}]{\footnotemark[#1]}

\title{Discourse-Aware Unsupervised Summarization of Long Scientific Documents} 

\author{%
  Yue Dong\thanks{\;\;Equal contribution.}\\
  MILA/McGill University \\
  Montreal, QC, Canada \\
  \texttt{yue.dong2} \\
  \texttt{@mail.mcgill.ca} \\
   \And
  Andrei Mircea\samethanks\\
  MILA/McGill University \\
  Montreal, QC, Canada \\
  \texttt{andrei.romascanu} \\
  \texttt{@mail.mcgill.ca} \\
   \And
Jackie C. K. Cheung\\
  MILA/McGill University \\
  Montreal, QC, Canada \\
  \texttt{jcheung} \\
  \texttt{@cs.mcgill.ca} \\
}

\date{}

\begin{document}
\maketitle
\begin{abstract}
We propose an unsupervised graph-based ranking model for  extractive summarization of long scientific documents. Our method assumes a two-level hierarchical graph representation of the source document, and exploits asymmetrical positional cues to determine sentence importance. Results on the PubMed and arXiv datasets show that our approach\footnote{Link to our code: \url{https://github.com/mirandrom/HipoRank}.} outperforms strong unsupervised baselines by wide margins in automatic metrics and human evaluation. In addition, it achieves performance comparable to many state-of-the-art supervised approaches which are trained on hundreds of thousands of examples. These results suggest that patterns in the discourse structure are a strong signal for determining importance in scientific articles.
\end{abstract}

\section{Introduction}\label{introduction}
\begin{table}[t!]
\resizebox{\columnwidth}{!}{
\begin{tabular}{  m{5em} |  m{7cm}  } 
\toprule
\textcolor[HTML]{0073b7}{Introduction} & anxiety affects quality of life in those living with parkinson's disease (pd) more so than overall cognitive status, motor deficits, apathy, and depression.\\ 
\midrule
 \textcolor[HTML]{0073b7}{Introduction} & although anxiety and depression are often related and coexist in pd patients, recent research suggests that anxiety rather than depression is the most prominent and prevalent mood disorder in pd.\\
\midrule
\textcolor[HTML]{f39c12}{Related Work} &furthermore, since previous work, albeit limited, has focused on the influence of symptom laterality on anxiety and cognition, we also explored this relationship .\\
\midrule
\textcolor[HTML]{e74c3c}{Methodology} & this study is the first to directly compare cognition between pd patients with and without anxiety. \\ 
\midrule
 \textcolor[HTML]{00a65a}{Result} & the findings confirmed our hypothesis that anxiety negatively influences attentional set-shifting and working memory in pd. \\ 
\midrule
\textcolor[HTML]{00a65a}{Result} &moreover, anxiety has been suggested to play a key role in freezing of gait (fog), which is also related to attentional set-shifting.\\
\midrule
\textcolor[HTML]{605ca8}{Future work} & s. future research should examine the link between anxiety, set-shifting, and fog, in order to determine whether treating anxiety might be a potential therapy for improving fog.\\ 
\bottomrule
\end{tabular}
}
\caption{Example of a PubMed article's summary produced by our model \textsc{HipoRank}. The  hierarchical and directed graph combined with  discourse-aware edge weighting allow \textsc{HipoRank} to generate summaries that   cover topics from different sections of the scientific article.}
\end{table}

Single document summarization aims at shortening a text and preserving the most important ideas of the source document. 
While abstractive strategies generate summaries with novel words,  extractive strategies select sentences from the source to form a summary \cite{nenkova2011automatic}. Despite recent advances in abstractive summarization, extractive models are still attractive in cases where faithfully preserving the original text is the priority. For example, legal arguments can hinge on the exact wording of a contract \cite{farzindar2004legal}, and ensuring the factual correctness of a summary can be critical in the health or scientific domains, which is a known weakness of current abstractive methods \cite{kryscinski2019evaluating}.

Supervised neural-based models have been the dominant paradigm in recent extractive systems, at least for \textit{short news summarization} \citep{nallapati2017summarunner,dong2018banditsum,zhou2018neural,liu2019text,narayan2018ranking,zhang2019hibert}. These models usually employ the encoder-decoder structure and have achieved promising performance on news datasets such as CNN/DailyMail \citep{hermann2015teaching}, and NYT \citep{sandhaus2008new}. 

However, these models cannot easily be adapted to out-of-domain data that have greater length and fewer training examples such as scientific article summarization \citep{xiao2019extractive} due to two significant limitations. First, they require large domain-specific training pairs of source documents and gold-standard summaries, which are often not available or feasible to create \citep{zheng2019sentence}. Second, the typical setup of using a token-level encoder-decoder with an attention mechanism does not scale well to longer documents \cite{shao2017generating}, as the number of attention computations is quadratic with respect to the number of tokens in the input document.

We instead explore \emph{unsupervised} approaches to address these challenges on long document summarization. We show that a simple unsupervised graph-based ranking model combined with proper sophisticated modelling of discourse information as an inductive bias can achieve unreasonable effectiveness in selecting important sentences from long scientific documents. 

For the choice of unsupervised graph-based ranking model, we follow the paradigm of LexRank \citep{erkan2004lexrank} and \textsc{PacSum} \citep{zheng2019sentence}. In these methods, sentences are nodes and weighted edges represent the degree of similarity between sentences. Summary generation is formulated as a node selection problem, in which nodes (i.e., sentences) that are semantically similar to other nodes are chosen to be included in the final summary. In other words, they determine node importance by defining a notion of centrality in the graph.

In addition, we augment the document graph with directionality and hierarchy to reflect the rich discourse structure of long scientific documents. In particular, our method relies on two insights about the discourse structure of long scientific documents. The first is that important information typically occurs at the start and end of sections; i.e., they tend to appear near section boundaries \citep{baxendale1958machine,lin1997identifying,teufel1997sentence}. We implement this using an asymmetric edge weighting function in a \textit{directed graph} which considers the distance of a sentence to a boundary. The second is that most sentences across section boundaries are unlikely to interact significantly with each other \citep{xiao2019extractive}. We implement this insight by injecting \textit{hierarchies} into our model, introducing section-level representations as graph nodes in addition to sentence nodes. By doing so, we convert a flat graph into a hierarchical non-fully-connected graph, which has two advantages: 1) reduced computational cost and 2) pruning of distracting weak connections between sentences across different sections.

We call our approach \textbf{Hi}erarchical and \textbf{Po}sitional \textbf{Rank}ing model (\textsc{HipoRank}) and evaluate it on summarizing long scientific articles from PubMed and arXiv \cite{cohan2018discourse}. Empirical results show that our method significantly improves performance over previous unsupervised models \cite{zheng2019sentence,erkan2004lexrank} in both automatic and human evaluation. In addition, our simple unsupervised approach achieves performance comparable to many expensive state-of-the-art supervised neural models that are trained  on hundreds of thousands of examples of long document pairs \cite{xiao2019extractive,subramanian2019extractive}. This suggests that patterns in the discourse structure are highly useful for determining sentence importance in long scientific articles, and that explicitly building in biases inspired by this structure is a viable strategy for building summarization systems.

\section{Related Work}
\subsection{Extractive Summarization}
Traditional extractive summarization methods are
mostly unsupervised \citep{radev2000centroid,lin2002single,wan-2008-exploration,wan2008multi,hirao-etal-2013-single,parveen2015topical,yin2015optimizing,li2017salience,zheng2019sentence}, utilizing a notion of sentence importance based on n-gram overlap with other sentences and frequency information 
\citep{nenkova2005impact}, relying on graph-based methods for sentence ranking \citep{erkan2004lexrank,mihalcea2004textrank}, or performing keyword extraction combined with submodular maximization \citep{tixier2017combining,shang2018unsupervised}.  

With the development of large-scale summarization datasets such as CNN/DailyMail \citep{hermann2015teaching}, NYT \citep{sandhaus2008new}, Newsroom \citep{grusky2018newsroom} and XSum \citep{narayan2018don}, along with advancements in deep neural-based architectures, modern supervised neural network-based methods that employ encoder-decoder framework have become increasingly popular. These models have been proposed with extractive strategies \citep{cheng-lapata-2016-neural,nallapati2017summarunner,wu2018learning,dong2018banditsum,zhou2018neural,narayan2018ranking}; abstractive strategies \citep{see2017get,chen2018fast,gehrmann2018bottom, dong2019unified,zhang2019pegasus,lewis2019bart}; and hybrid strategies \citep{hsu-etal-2018-unified, bae2019summary,moroshko2019editorial}.

More recently, extractive approaches leveraging transformer architectures \citep{vaswani2017attention} and their pretrained counterparts \citep{devlin2019bert,lewis2019bart,zhang2019pegasus,dong2019unified} have achieved state-of-the-art performances on the CNN/DailyMail news benchmark dataset \citep{zhang2019hibert,liu2019text,zhong2019searching}. Furthermore, pretrained transformer models also provide better sentence representations for unsupervised summarization methods. For instance, \textsc{PacSum} \citep{zheng2019sentence}, a directed graph-based unsupervised model that utilizes BERT-based sentence representations, achieved comparable performance to supervised models on the CNN/DailyMail and NYT datasets.  

\subsection{Extractive Summarization of Long Scientific Papers}
Despite the success of deep neural-based models on news summarization, these approaches typically face challenges when applied to long documents such as scientific articles. 
Furthermore, these approaches are often blind to the topical information resulting from the structured sections in scientific articles \citep{xiao2019extractive}. Two recent neural supervised models address these issues. \citet{subramanian2019extractive} used the introduction section as a proxy for the whole document, while \citet{xiao2019extractive} divided articles into sections and used non-auto-regressive approaches to model global and local information. 

Besides neural approaches, most previous scientific article summarization systems employ traditional supervised machine learning algorithms with surface features as input \citep{xiao2019extractive}. Surface features such as sentence position, sentence and document length, keyphrase score, and fine-grain rhetorical categories are often combined with Naive Bayes \citep{teufel2002summarizing}, CRFs
and SVMs \citep{liakata-etal-2013-discourse}, LSTM and MLP \citep{collins2017supervised} for extractive summarization over long scientific articles. To the best of our knowledge, the only unsupervised extractive summarization model for long scientific documents relies on citation networks \citep{qazvinian-2008-citation-network, cohan-goharian-2015-scientific}, by extracting citation-contexts from citing articles and ranking these sentences to form the final summary. Our proposed method is different from their settings, where we perform single document summarization based on the long source article.

\section{Method}
\begin{figure}[t]
    \centering
    \includegraphics[width=\columnwidth]{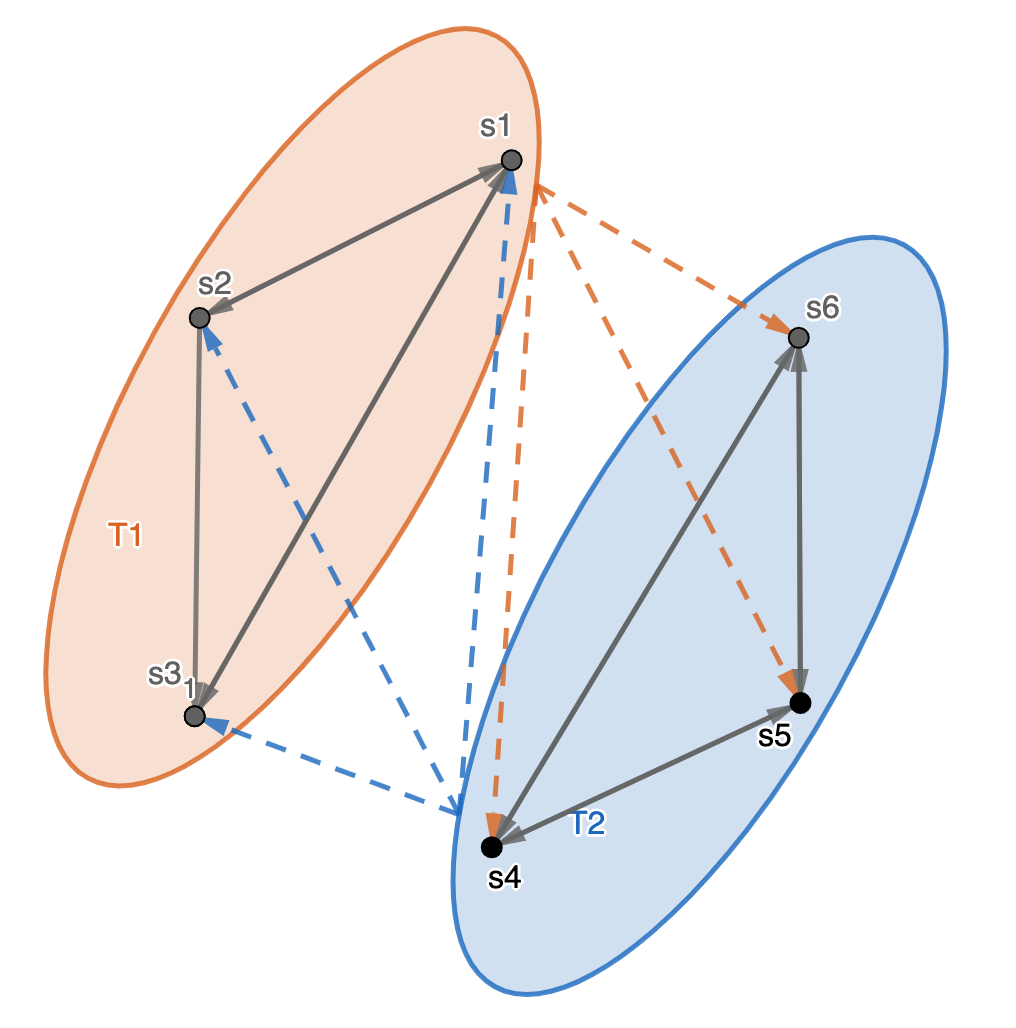}
    \caption{Example of a hierarchical document graph constructed by our approach on a toy document that contains two sections $\{T_1, T_2\}$, each containing three sentences for a total of six sentences $\{s_1, \ldots,\ s_6\}.$ Each double-headed arrow represents two edges with opposite directions. The solid and dashed arrows indicate intra-section and inter-section connections respectively. When compared to the flat fully-connected graph of traditional methods, our use of hierarchy effectively reduces the number of edges from 60 to 24 in this example. 
    }
    \label{fig:hierarchy}
\end{figure}

Our proposed method combines simple graph-based ranking algorithms with a two-level hierarchical model of the rich discourse structures of long scientific documents \citep{teufel1997sentence,xiao2019extractive}. We incorporate this discourse information into the graph as inductive biases through the construction of a \textit{directed hierarchical graph} for document representation (Figure \ref{fig:hierarchy} and Section \ref{sub-sec:method_hierarchy}) and through the asymmetric edge weighting of edges with boundary functions (Section \ref{sub-sec:method_edgeweights}). 

\subsection{Graph-based Ranking Algorithm}
Graph-based ranking algorithms for summarization represent a document as a graph $G = (V, E)$, where $V$ is the set of vertices that represent sentences or other textual units in the document, and $E$ is the set of edges that represent interactions between sentences. The directed edge $e_{ij}$ from node $v_i$ to node $v_j$ is typically weighted by $w_{ij}=f(sim(v_i,v_j))$, where $sim$ is a measure of similarity between two nodes (e.g. cosine distance between their distributed representations), and $f$ can be an additional weighting function. These algorithms select the most salient sentences from $V$ based on the assumption that sentences that are similar to a greater number of other sentences capture more important content and therefore are more informative.

\subsection{Hierarchical Document Graph Creation}\label{sub-sec:method_hierarchy}
To create a hierarchical document graph, we first split a document into its sections, then into sentences\footnote{Our approach is agnostic to the sentence/section splitting method. In our experiments, articles in the datasets are already split into sections and sentences.}.  To create the hierarchy, we allow two levels of connections in our hierarchical graph: intra-sectional connections and inter-sectional connections as shown in Figure \ref{fig:hierarchy}.
\paragraph{Intra-sectional connections} aim to model the \textit{local} importance of a sentence within its section. It implements the idea that a sentence that is similar to a greater number of other sentences in the same topic/section should be more important. This is realized in our fully-connected subgraph for an arbitrary section $I$, where we allow  \textit{sentence-sentence} edges for all sentence nodes within the same section.
\paragraph{Inter-sectional connections}
aim to model the \textit{global} importance of a sentence with respect to other topics/sections in the document, as a sentence that is similar to a greater number of other topics is deemed more important. However, calculating sentence-sentence connections across different sections is computationally expensive and may also suffer from performance degradation due to weak edges between sentences that are unrelated as a result of being from different sections \citep{mihalcea2004textrank}. To address these issues, We introduce section nodes on top of sentence nodes to form a hierarchical graph. For inter-section connections, we only allow \textit{section-sentence} edges for modeling the global information. This choice makes our approach more computationally efficient while greatly limiting the number of irrelevant inter-section edges that arise from the fact that sections in scientific documents typically have independent topics \citep{xiao2019extractive}.  In contrast, traditional graph-based ranking algorithms have a flat fully-connected graph document with no sections.

\subsection{Asymmetric Edge Weighting by Boundary functions}\label{sub-sec:method_edgeweights}
To calculate the weight of an edge, we first measure similarity between a sentence-sentence pair  $sim({v^I_{j}, v^I_{i}})$ and a section-sentence pair $sim(v^J,v^I_i)$. While our method is agnostic to the measure of similarity, we use cosine similarity with different vector representations in our experiments, averaging a section's sentence representations to obtain its own. 

While the similarities of two graph nodes are symmetric, one may be more salient than the other when considering  their discourse structures \citep{baxendale1958machine,teufel1997sentence}.  Based on these discourse hypotheses of long scientific documents, we capture this asymmetry by making
our hierarchical graph \textit{directed} and inject \textit{asymmetric} edge weighting over intra-section and inter-section connections. 

\paragraph{Asymmetric edge weighting over sentences}
Our asymmetric edge weighting is based on the hypothesis that important sentences are near the boundaries (start or end) of a text \citep{baxendale1958machine}.  We reflect this hypothesis by defining a \textit{sentence boundary function} $d_b$ over sentences $v^I_i$ in section $I$ such that sentences closer to the section's boundaries are more important:
\begin{equation}
\label{eq:sent_boundary_func}
d_b(v^I_i) = \min(x^I_i,\alpha(n^I-x^I_i)),
\end{equation}
where $n^I$ is the number of sentences in section $I$  and $x^I_i$ represents sentence $i$'s position in the section $I$.  $\alpha \in \mathbb{R}^+$ is a hyper-parameter that controls the relative importance of the start or end of a section or document.   

The sentence boundary function allow us to incorporate directionality in our edges, and weight edges differently depending on if they are incident to a more important or less important sentence in the same section. Concretely, we define the weight $w^I_{ji}$ for intra-section edges (incoming edges for $i$) as:
\begin{equation}\label{eq:sent_boundary_cond}
        w^I_{ji}= 
    \begin{cases}
    \lambda_1*sim(v^I_j, v^I_i), & \textrm{if} \: d_b(v^I_i) \geq d_b(v^I_j),\\
    
    \lambda_2*sim(v^I_j,v^I_i), & \textrm{if} \: d_b(v^I_i) < d_b(v^I_j)\\
    \end{cases}
    \end{equation}
where $\lambda_1 <\lambda_2$ such that an edge $e_{ji}$ incident to $i$ is weighted more if $i$ is closer to the text boundary than $j$. Edges with a weight below a certain threshold $\beta$ can be pruned (i.e., set to $0$). 
    
\paragraph{Asymmetric edge weighting over sections}
Similarly, to reflect the hierarchy hypothesis over long scientific documents proposed by \citet{teufel1997sentence}, we also define a \textit{section boundary function} $d_b$ to reflect that sections near a document's boundaries are more important:
\begin{equation}
\label{eq:sec_boundary_func}
d_b(v^I) = \min(x^I,\alpha(N-x^I)),
\end{equation}
where $N$ is the number of sections in the document  and $x^I$ represents section $I$'s position in the document.  

This section boundary function allows us to inject asymmetric edge weighting $w^{JI}_{i}$ to inter-section edges:
    \begin{equation}\label{eq:sec_boundary_cond}
        w^{JI}_{i}= 
    \begin{cases}
    \lambda_1*sim(v^J, v^I_i), & \textrm{if} \: d_b(v^I) \geq d_b(v^J).\\
    \lambda_2*sim(v^J, v^I_i), & \textrm{if} \: d_b(v^I) < d_b(v^J)\\
    \end{cases}
    \end{equation}
where $\lambda_1 <\lambda_2$ such that an edge $e^{JI}_{i}$ incident to $i \in I$ is weighted more if section $I$ is closer to the text boundary than section $J$. 
    
\subsection{Importance Calculation}\label{sub-sec:method_centrality}
We compute the overall importance of sentence $v^I_i$ as the weighted sum of its inter-section and intra-section centrality scores:
  \begin{equation}\label{eq:importance}
        c(v^I_i) = \mu_1 \cdot c_{\text{inter}}(v^I_i) + c_{\text{intra}}(v^I_i)
    \end{equation}
    
  \begin{equation}\label{eq:centrality}
    \begin{split}
    c_{\text{intra}}(v^I_i) &= \sum_{v^I_j \in I} \dfrac{ w^I_{ji}}{|I|} \\  
    c_{\text{inter}}(v^I_i) &= \sum_{v^J \in D} \dfrac{ w^{JI}_{i}}{|D|}, \\  
    \end{split}
    \end{equation}
where $I$ is the set of sentences neighbouring $v^I_i$ and $D$ is the set of neighbouring sections in the hierarchical document graph; $\mu_1$ is a weighting factor for  inter-section centrality.

\subsection{Summary Generation}\label{sub-sec:method_generation}
Lastly, we generate a summary by greedily extracting sentences with the highest importance scores until a predefined word-limit $L$ is passed. Most graph-based ranking algorithms recompute importance after each sentence is extracted in order to prevent content overlap. However, we find that the asymmetric edge scoring functions in (\ref{eq:sent_boundary_cond}) and (\ref{eq:sec_boundary_cond}) naturally prevent redundancy, because similar sentences have different boundary positional scores. Our method thus successfully extracts diverse sentences without recomputing importance.

\section{Experimental Setup}
This section describes the datasets, the hyperparameter choices, the baseline models, and the evaluation metrics used in the experiments. 

\subsection{Datasets}
Our experiments are conducted on PubMed and arXiv \cite{cohan2018discourse}, two large-scale datasets of long and structured scientific articles with abstracts as summaries. The average source article length is four to seven times longer than popular news benchmarks (Table \ref{tab:datasets}), making them ideal candidates to test our method. 
\begin{table}[t!]
\small
\centering
\begin{tabular}{|l|l|c|c|}
\toprule
Dataset & \# docs & avg. doc. len. & avg. summ. len. \\ \midrule
CNN & 92K & 656 & 43 \\ 
Daily Mail & 219K & 693 & 52 \\ 
NYT & 655K & 530 & 38 \\ 
PubMed & 133K & 3,016 & 203 \\ 
arXiv & 215K & 4,938 & 220 \\ 
\bottomrule
\end{tabular}
\caption{\small Dataset statistics on news articles (CNN, DailyMail, and NYT) and long scientific documents (PubMed and arXiv).} \label{tab:datasets}
\end{table}

\subsection{Implementation Details}\label{sub-sec:hyperparams}
Our model's hyperparameters for testing are chosen from the ablation studies on the validation sets. The test results are reported with the following hyperparameter settings: $\lambda_1=0.0, \lambda_2=1.0, \alpha=1.0$, with $\mu_1 = 0.5$ for PubMed and $\mu_1 = 1.0$ for arXiv.
We fix $\lambda_2$ to 1 and the choices of $\lambda_1 \in \{-0.2, 0, 0.5\}$. represent whether the edge between a less boundary-important sentence and a more boundary-important sentence is 1) negatively weighted, 2) pruned, or 3) down-weighted. $\lambda_1 < \lambda_2$ such that an edge $e_{ji}$ incident to $i$ is weighted more if $i$ is closer to the text boundary than $j$. $\alpha \in \{0, 0.5, 0.8, 1.0, 1.2\}$ controls the relative importance of the start or end of a section or document. $\mu_1 \in \{0.5, 1.0, 1.5\}$ controls how much we weigh intra-section sentence importance vs. inter-section sectional importance.

For each dataset, we experimented with different pretrained distributional sentence representation models. The dimension of sentence representations is model-dependent (details in Section \ref{sub-sec:embeddings}).  We used the publicly released BERT model\footnote{https://github.com/huggingface/transformers} \citep{devlin2019bert}, \textsc{PacSum} BERT model\footnote{https://github.com/mswellhao/PACSUM} \citep{zheng2019sentence}, SentBERT and SentRoBERTa\footnote{https://github.com/UKPLab/sentence-transformers} \citep{reimerssentence}, and BioMed word2vec representations\footnote{http://bio.nlplab.org/word-vectors} \citep{moen2013distributional}. A section's representation is calculated as the average of its sentences' representations. The similarity between sentences or sections is defined to be the cosine similarity between the distributed representations. 

\subsection{Baselines}\label{sub-sec:baseline_models}
We compare our approach with previous unsupervised and supervised models in extractive summarization. In addition, we also compare it with recent neural abstractive approaches for completeness. 

For unsupervised extractive summarization models, we compare with SumBasic \citep{vanderwende2007beyond},  LSA \citep{steinberger2004using}, LexRank \citep{erkan2004lexrank} and \textsc{PacSum} \citep{zheng2019sentence}. For supervised neural extractive summarization models, we compare with a vanilla encoder-decoder model \citep{cheng-lapata-2016-neural}, SummaRuNNer \citep{nallapati2017summarunner}, GlobalLocalCont \citep{xiao2019extractive}, Sent-CLF and Sent-PTR \citep{subramanian2019extractive}. We also compare with neural abstractive summarization models as reported in \citet{xiao2019extractive}: Attn-Seq2Seq \citep{nallapati2016abstractive}, Pntr-Gen-Seq2Seq \citep{see2017get} and Discourse-aware \citep{cohan2018discourse}. In addition, we report the lead baseline that selects the first $k$ tokens as a summary ($k=203,=220$ for PubMed and arXiv respectively). Lastly, we report baselines for an Oracle summarizer \citep{nallapati2017summarunner}. 

\subsection{Evaluation Methods}\label{sub-sec:evalution_method}
We evaluate our method with automatic evaluation metrics - ROUGE F1 scores \citep{lin-2004-rouge}. ROUGE-1 and ROUGE-2 compute unigram and bigram overlaps between system summaries and reference summaries, while ROUGE-L computes the longest common sub-sequence of the two. 

In addition, we design a human evaluation experiment (details in Section \ref{sub-sec:human_eval_result}) to compare our model with the best unsupervised summarization model - \textsc{PacSum} \citep{zheng2019sentence}. As far as we know, we are the first to perform human evaluation on the 2018 PubMed and arXiv datasets \citep{cohan2018discourse}.  Human evaluation over long scientific articles require annotators to comprehend a full domain-specific long article and compare multiple summaries for quality evaluation. Due to the challenging nature of the task,  previous papers choose to skip it and purely rely on automatic evaluations to judge the system performance.

\section{Results}
\subsection{Automatic Evaluation Results}\label{sub-sec:automatic_eval_result}
Tables~\ref{table:pubmed-test-table} and \ref{table:arxiv-test-table} summarize our automatic evaluation results on the PubMed and arXiv test sets  with the best hyperparameters, as described in Section \ref{sub-sec:hyperparams}.
    \begin{table}[t!]
    \centering
    \resizebox{\columnwidth}{!}{%
    \begin{tabular}{l c c c }
    \toprule
     \textbf{\scriptsize Model} & \textbf{\scriptsize ROUGE-1} &  \textbf{\scriptsize ROUGE-2} & \textbf{\scriptsize ROUGE-L}\\ 
    \midrule
    Lead &35.63& 12.28& 25.17 \\
    Oracle (ROUGE-2, F1) &55.05 &27.48& 38.66 \\
    \midrule
    \multicolumn{4}{c}{Supervised Abstractive}\\
    \midrule
    Attn-Seq2Seq (2016) &31.55& 8.52& 27.38\\
    Pntr-Gen-Seq2Seq (2017) &35.86 &10.22& 29.69 \\
    Discourse-aware (2018) &38.93 &15.37 &35.21\\
    \midrule
    \multicolumn{4}{c}{Supervised Extractive}\\
    \midrule
    Cheng \& Lapata (2016)& 43.89 &18.53 &30.17 \\
    SummaRuNNer (2017) &43.89 &18.78 &30.36 \\
    GlobalLocalCont (2019) & 44.85 &19.70 &31.43 \\
    Sent-{CLF} (2019)& 45.01 & 19.91& 41.16\\
    Sent-PTR (2019) & 43.30 & 17.92 & 39.47 \\
    \midrule
    
    \multicolumn{4}{c}{Unsupervised Extractive}\\
    \midrule
    SumBasic (2007) &37.15&11.36& 33.43\\
    LSA (2004) &33.89 &9.93 &29.70\\
    LexRank (2004) &39.19 &13.89& 34.59\\
    \textsc{PacSum} (2019) & 39.79 & 14.00 & 36.09  \\
    \textsc{HipoRank} (ours) & \textbf{43.58} & \textbf{17.00} & \textbf{39.31} \\
    \bottomrule%
    \end{tabular}
    }
    \caption{\label{table:pubmed-test-table}Test set results on PubMed (ROUGE F1). }
    \hfill
\end{table}

    \begin{figure*}[ht]
      \subfigure[Oracle]{\includegraphics[width=.32\textwidth]{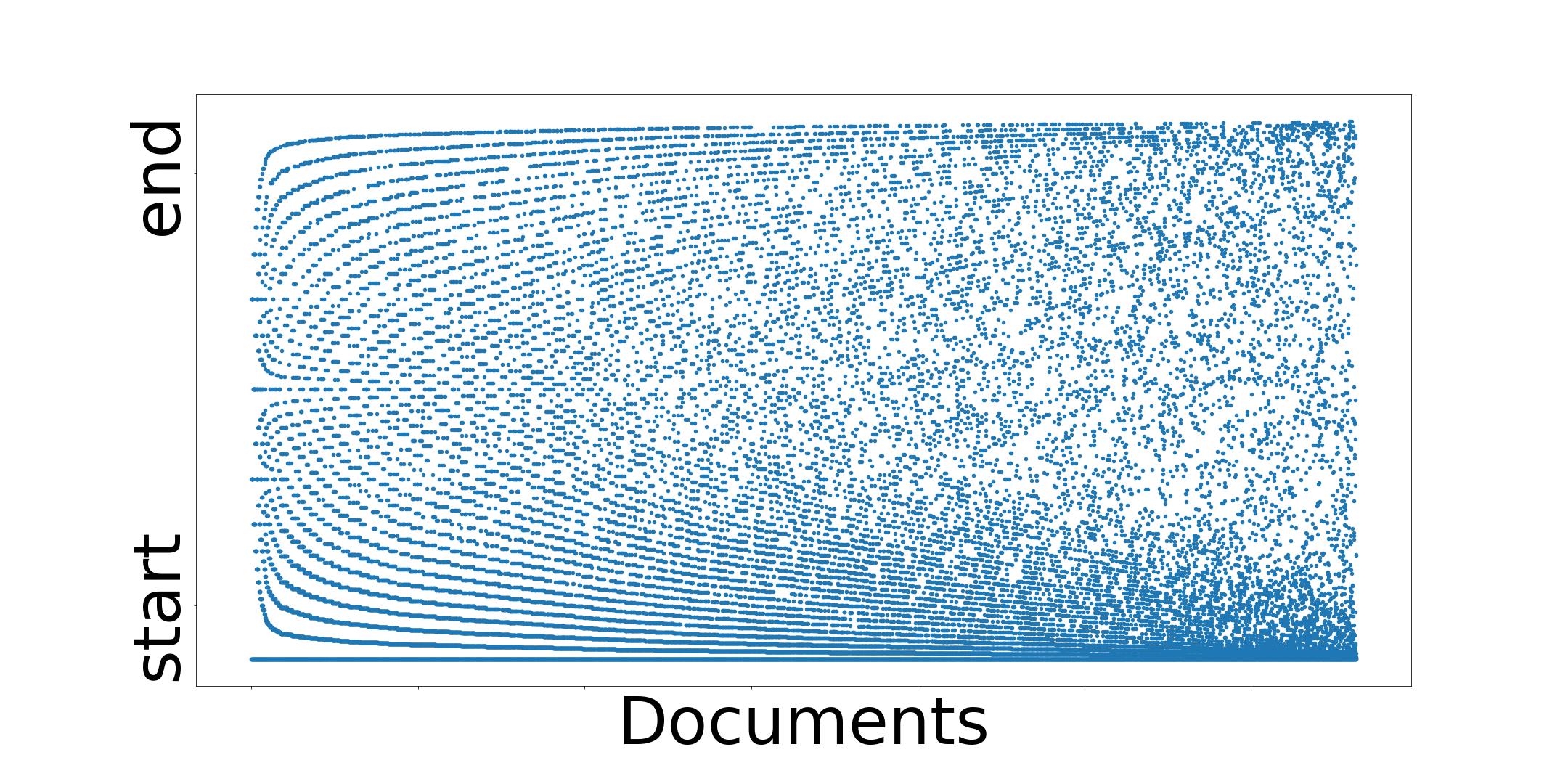} \label{sub-fig:pubmed1}} \hfill
      \subfigure[\textsc{PacSum}]{\includegraphics[width=.32\textwidth]{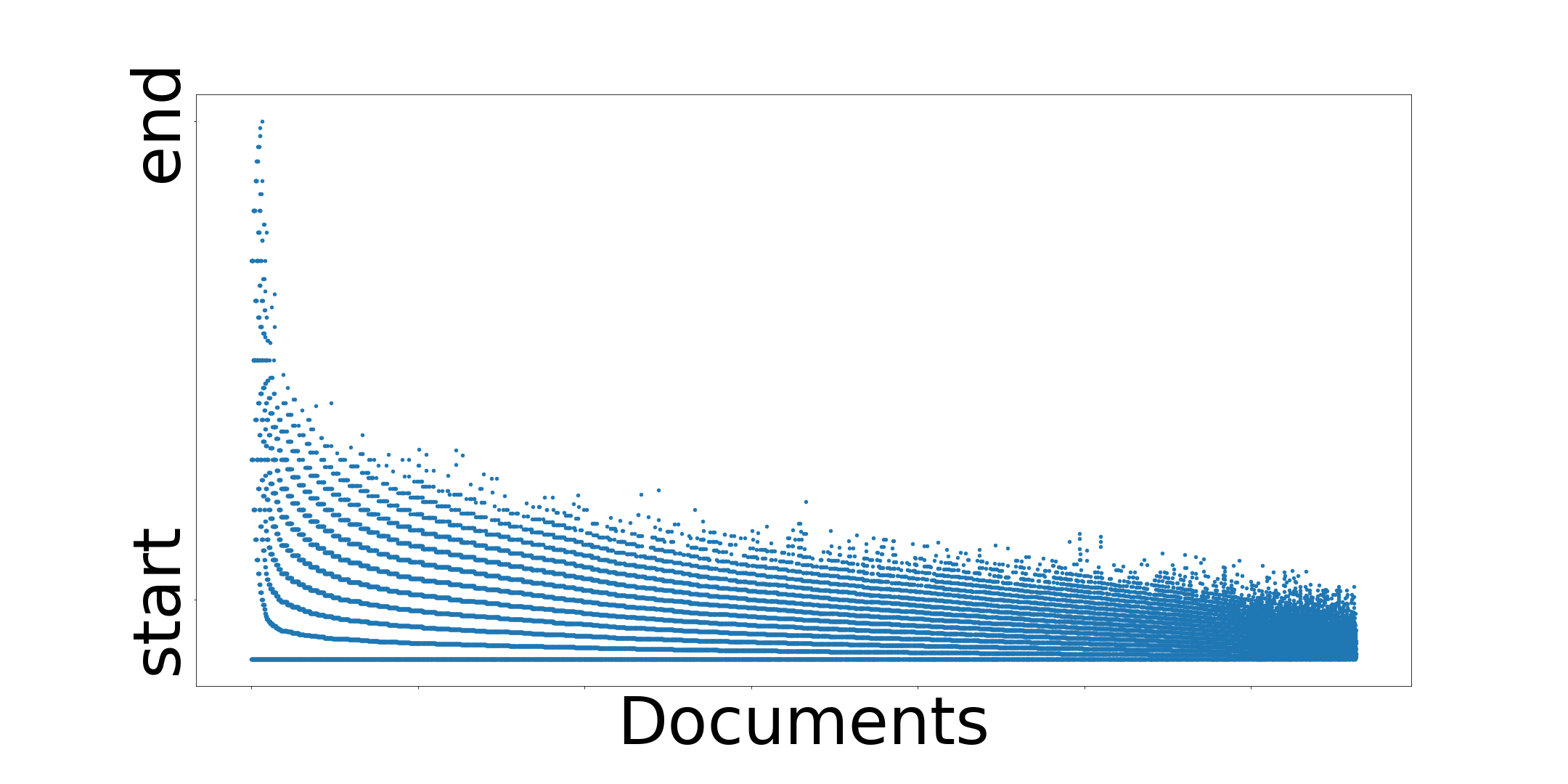} \label{sub-fig:pubmed2}} \hfill
      \subfigure[\textsc{HipoRank}]{\includegraphics[width=.32\textwidth]{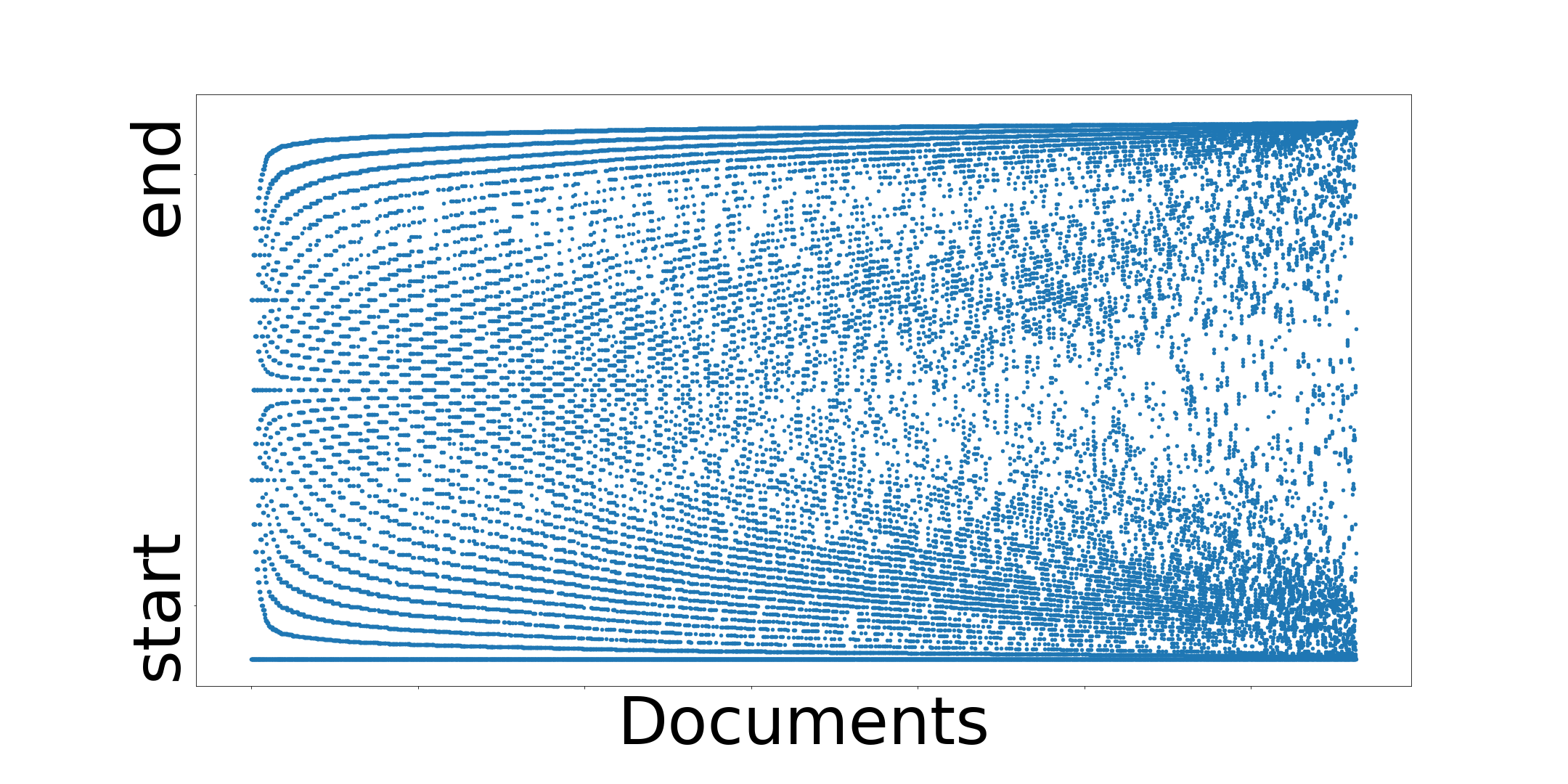} \label{sub-fig:pubmed3}} \hfill
    
    \caption{Sentence positions in source document for extractive summaries generated by different models on the PubMed validation set. Documents on the x-axis are ordered by increasing article length from shortest to longest. We also see a similar trend on arXiv (the plots with more details can be found in the appendix).}
    \label{fig:oracle_density_plot}
    \end{figure*}
    
\begin{table}[t!]
    \centering
    \resizebox{\columnwidth}{!}{
    \begin{tabular}{l c c c }
    \toprule
     \textbf{\scriptsize Model} & \textbf{\scriptsize ROUGE-1} &  \textbf{\scriptsize ROUGE-2} & \textbf{\scriptsize ROUGE-L} \\ 
    \midrule
    Lead &33.66 &8.94 &22.19 \\
    Oracle (ROUGE-2, F1) &53.88& 23.05 &34.90 \\
    \midrule
    \multicolumn{4}{c}{Supervised Abstractive}\\
    \midrule
    Attn-Seq2Seq (2016)& 29.30 &6.00& 25.56\\
    Pntr-Gen-Seq2Seq (2017) & 32.06& 9.04 &25.16\\
    Discourse-aware (2018) & 35.80 &11.05& 31.80 \\
    \midrule
    \multicolumn{4}{c}{Supervised Extractive}\\
    \midrule
    Cheng\&Lapata (2016) &42.24& 15.97& 27.88\\
     SummaRuNNer (2017) &42.81 &16.52 &28.23 \\
    GlobalLocalCont (2019) &43.62 &17.36& 29.14 \\
    Sent-{CLF} (2019) & 34.01& 8.71 & 30.41\\
    Sent-PTR (2019) & 42.32 &15.63 & 38.06\\
     \midrule
    \multicolumn{4}{c}{Unsupervised Extractive}\\
    \midrule
    SumBasic (2007) &29.47 &6.95& 26.30 \\
    LSA (2004) & 29.91 &7.42 & 25.67 \\
    LexRank (2004) & 33.85 &10.73 & 28.99 \\
    \textsc{PacSum} (2019) & 38.57 & 10.93 & 34.33  \\
    \textsc{HipoRank} (ours) & \textbf{39.34} & \textbf{12.56} & \textbf{34.89} \\
    \bottomrule
    \end{tabular}
    }
    \caption{\label{table:arxiv-test-table} Test set results on arXiv (ROUGE F1).}
    \hfill
\end{table}

The first blocks in Table \ref{table:pubmed-test-table},\ref{table:arxiv-test-table} include the lead  and the oracle baselines. The second and the third blocks in the tables present the results of supervised abstractive models, and of supervised extractive models. ROUGE-2 oracle summaries are used as gold standard summaries for training supervised extractive models, which likely contributes to their better ROUGE-2 scores.

The last blocks compare previous unsupervised models with our approach. Our model outperforms all other unsupervised approaches by wide margins in terms of ROUGE-1,2,L F1 scores on both PubMed and arXiv datasets. We also show that \textsc{PacSum} is biased towards selecting sentences that appear at the beginning of a document while our method selects sentences in every section and near the article boundaries, similar to the oracle (Figure \ref{fig:oracle_density_plot}). This overlap with gold standard summaries suggests our use of discourse structure and hierarchy plays a significant role in our method's performance. 

Interestingly, despite limited access to only the validation set for hyperparameter tuning, our method achieves performance scores that are competitive with supervised models that require hundreds of thousands of training examples, outperforming almost all abstractive and extractive models on ROUGE-L. This suggests that our discourse-aware unsupervised model is surprisingly effective at selecting salient sentences in long scientific document and perhaps should be used as a strong baseline to accessing the merits of supervised approaches for learning content beyond discourse. 
 
\subsection{Human Evaluation}\label{sub-sec:human_eval_result}
    \begin{table}[!t]
    \centering
    \resizebox{\columnwidth}{!}{%
    \begin{tabular}{l c c}
    \toprule
     Model & Content-coverage &   Importance \\ 
    \midrule
    \textsc{PacSum} &  30.52 & 48.70 \\ 
    \textsc{HipoRank} (ours) & \textbf{42.13} & \textbf{59.06} \\ 
    \bottomrule
    \end{tabular}
    }
\caption{\label{table:pubmed-human_eval_results} Human evaluation results on 20 sampled reference summaries with 281 system summary sentences from PubMed. Each reference summary-sentence pair is annotated by two annotators with an average annotator agreement of 73.24\%. The results are averaged across 127 sentences from HipoRank and154 sentences from state-of-the-art unsupervised extractive summarization system \textsc{PacSum} \citep{zheng2019sentence}..}\label{tab:human_eval}
\end{table}

We asked the human judges\footnote{All judges are native English speakers with at least a bachelor's degree and experience in scientific research. We compensated the judges at an hourly rate of \$20.} to read the reference summary\footnote{We made the decision to not present the whole article, which would create a large cognitive burden on judges and incentivize them to take shortcuts.} (abstract) and present extracted sentences from different summarization systems in a random and anonymized order. The judges are asked to evaluate the system summary sentence according to two criteria
: 1) \textit{content coverage} (whether the presented sentence contains content from the abstract); and 2) \textit{importance} (whether the presented sentence is important for a goal-oriented reader even if it isn't in the abstract \citep{lin1997identifying}).

Table~\ref{tab:human_eval} presents the human evaluation results. 
\textsc{HipoRank} is shown to be significantly better than \textsc{PacSum} in  both content coverage and importance ($p=0.002$ and $p=0.007$ with Mann-Whitney U tests, respectively).
We also measure inter-rater reliability using Fleiss' $\kappa$ ($46.56$ for \textit{content-coverage} and $41.37$ for \textit{importance}). These results help support that our method's use of hierarchy and discourse structure  improves summarization quality.

\section{Ablation Studies}\label{sec:ablation}
\begin{table}[t!]
\centering
\resizebox{\columnwidth}{!}{%
\begin{tabular}{l c c c c}
\toprule
 \textbf{\scriptsize Model} & \textbf{\scriptsize ROUGE-1} &  \textbf{\scriptsize ROUGE-2} & \textbf{\scriptsize ROUGE-L} \\ 

\midrule
\multicolumn{4}{c}{\textsc{HipoRank} + Different Positional Functions}\\
\midrule
lead& 37.43 &12.13& 33.68\\
undirected & 40.66& 13.41 &36.55\\
boundary-distance (ours) & \textbf{43.42} & \textbf{16.76} & \textbf{39.23}  \\

\midrule
\multicolumn{4}{c}{\textsc{HipoRank} + Different Hierarchical Functions}\\
\midrule
w.o. hierarchy & 41.88& 15.39 &37.91\\
w. hierarchy (ours) & \textbf{43.42} & \textbf{16.76} & \textbf{39.23}  \\
\bottomrule%
\end{tabular}
}
\caption{\label{tab:ablation} Results on the PubMed validation set with different positional function or hierarchical information.}
\end{table}
    
\subsection{Component-wise Analysis} Table \ref{tab:ablation} presents the ablation study to assess the relative contributions of the boundary function and the hierarchical information. We keep all the hyperparameters unchanged with respect to the best setting in Section \ref{sub-sec:hyperparams} and either vary the positional function or the hierarchical structures.  We also found that the improvement of each components are stable across all the hyperparameters we tested (more details in the appendix). 

The first block of Table \ref{tab:ablation} reports the ablation results with different positional functions: no positional function \citep{erkan2004lexrank,mihalcea2004textrank}, lead bias function  \citep{zheng2019sentence}, and our proposed boundary function. We can see that using the wrong positional function hurts the model's performance when comparing no positional function with lead bias function. Our boundary positional function outperforms the lead or no positional functions significantly. 

The second block of Table \ref{tab:ablation} reports the results with or without the hierarchical structure. We observe that adding the hierarchical information results in a huge performance improvement. 

\subsection{Effect of Embeddings}\label{sub-sec:embeddings}
To disentangle the effect of sentence representation, we show PubMed test set results of our best model with different sentence embeddings in Table \ref{table:pubmed-embedding_results}. While pretrained transformer models finetuned on sentence similarity improve performance, \textsc{HipoRank} still consistently outperforms previous state-of-the-art unsupervised models (Table~\ref{table:pubmed-test-table}) even with random embeddings. These results once again suggest that our method's improvement can indeed be attributed to the use of hierarchy and discourse structure, rather than to the the choice of representations. 

    \begin{table}[t]
    \centering
    \resizebox{\columnwidth}{!}{%
    \begin{tabular}{l c c c }
    \toprule
     \textbf{\scriptsize Model} & \textbf{\scriptsize ROUGE-1} &  \textbf{\scriptsize ROUGE-2} & \textbf{\scriptsize ROUGE-L}\\  
    \midrule
    Lead &35.63& 12.28& 25.17 \\
    Oracle &55.05 &27.48& 38.66 \\
    \midrule
    
    \multicolumn{4}{c}{\textsc{HipoRank} with Different Embeddings}\\
    \midrule
    Random Embedding (d=200) & 43.05 & 16.69&38.63\\
    Biomed-w2v (d=200) & 43.70 &17.06 & 39.19  \\ 
    BERT (d=768) & 42.91 & 16.27 & 38.52  \\
    \textsc{PacSum-BERT} (d=768) & 43.58 & 17.00 & \textbf{39.31} \\
    SentBERT (d=768) & \textbf{43.59} & \textbf{17.08} & 39.07  \\
    SentRoBERTa (d=1024) & 43.55 & 17.06 & 39.07  \\
    \bottomrule%
    \end{tabular}
    }
    \caption{\label{table:pubmed-embedding_results}  PubMed test set results with \textsc{HipoRank} framework and different pretrained sentence and section embeddings.}
    \hfill
    \end{table}
    
\subsection{Stability of  Hyperparameters}
    \begin{figure}[h!]
        \centering
        \includegraphics[width=\columnwidth]{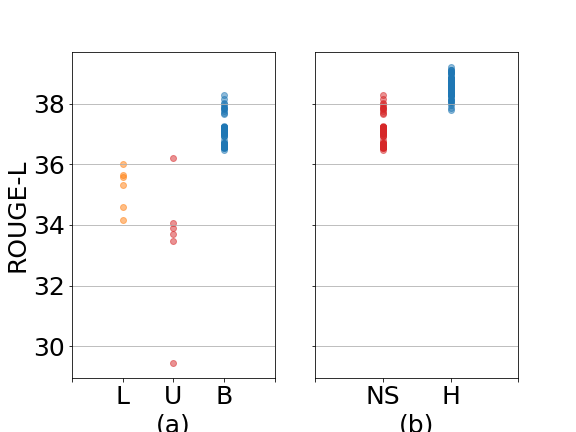}
        \caption{ROUGE-L scores for (a) different positional functions (L=lead, U=undirected, B=boundary) and (b) different graph hierarchies (NS=no\_section, H=hierarchical). Each point corresponds to one configuration of the hyperparameter gridsearch described in Section 4.2. }
        \label{fig:hyperparams}
    \end{figure}
    
To further inspect our model's stability across different hyperparameter choices, we conducted fine-grained analysis across all different hyperparameter settings as below.

\paragraph{Stability w.r.t. Discourse Structure}
To evaluate the impact and the stability of discourse structure informed edge weighting (Section 3.3), we first compared our \textit{boundary} positional function (Eqn. 1,3) to \textsc{PacSum}'s \textit{lead} positional function, as well as the standard \textit{undirected} approach over different hyperparameter settings.  Figure \ref{fig:hyperparams} (a) shows that our method consistently performed better on the PubMed validation set, across \textit{different hyperparameters and embedding models} outlined in Section 4.2.  

\paragraph{Stability w.r.t. Hierarchy}
We then evaluated the effect of adding hierarchy (Section 3.2) on top of our boundary positional function. In addition to decreasing the computational cost, Figure \ref{fig:hyperparams} (b) shows that incorporating hierarchy further improved ROUGE-L consistently across \textit{different hyperparameters and embedding models} we tested. 
    
\paragraph{Application to other genres}
While our work here is focused on long scientific document summarization, we believe that our approach is promising for other genres of text, provided that the right discourse-aware biases are given to the model. Indeed, one version of our model with our proposed boundary function can be seen as a generalization of \textsc{PacSum}, which achieves state-of-the-art performance on unsupervised summarization of news by exploiting the well known lead bias of news text \citep{zheng2019sentence,grenander-etal-2019-countering}. We leave such explorations of adapting \textsc{HipoRank} to other genres to future work.

\section{Conclusion}
We presented an unsupervised graph-based model for long scientific document summarization. The proposed approach augments the measure of sentence centrality by inserting directionality and hierarchy in the graph with boundary positional functions and hierarchical topic information grounded in discourse structure.
Our simple unsupervised approach with rich discourse modelling outperforms previous unsupervised graph-based summarization models by wide margins and achieves comparable performance to state-of-the-art supervised neural models. This makes our model a lightweight but strong baseline for assessing the performance of expensive supervised approaches for long scientific document summarization.

\section*{Acknowledgments}
This work is supported by the Natural Sciences and Engineering Research Council of Canada, Compute Canada, and the CIFAR Canada AI Chair program. We would like to thank Hao Zheng, Wen  Xiao, and Sandeep Subramanian for useful discussions. 

\bibliography{anthology,eacl2021}
\bibliographystyle{acl_natbib}

\clearpage
\newpage
\appendix
\section{Appendices}
\label{sec:appendix}
\begin{figure}[hbt!]
  \centering
   \subfigure[Flat fully-connected graph ]{\includegraphics[width=.35\textwidth]{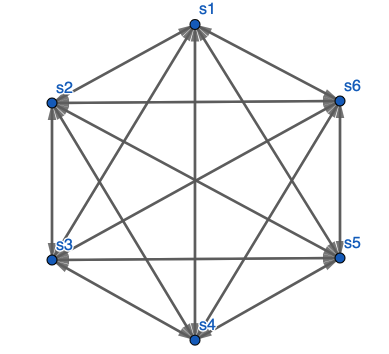} \label{sub-fig:hierarchy1}} \hfill
  \subfigure[Section-section hierarchical multiplication (hierarchy-multiply, ours)]{\includegraphics[width=.35\textwidth]{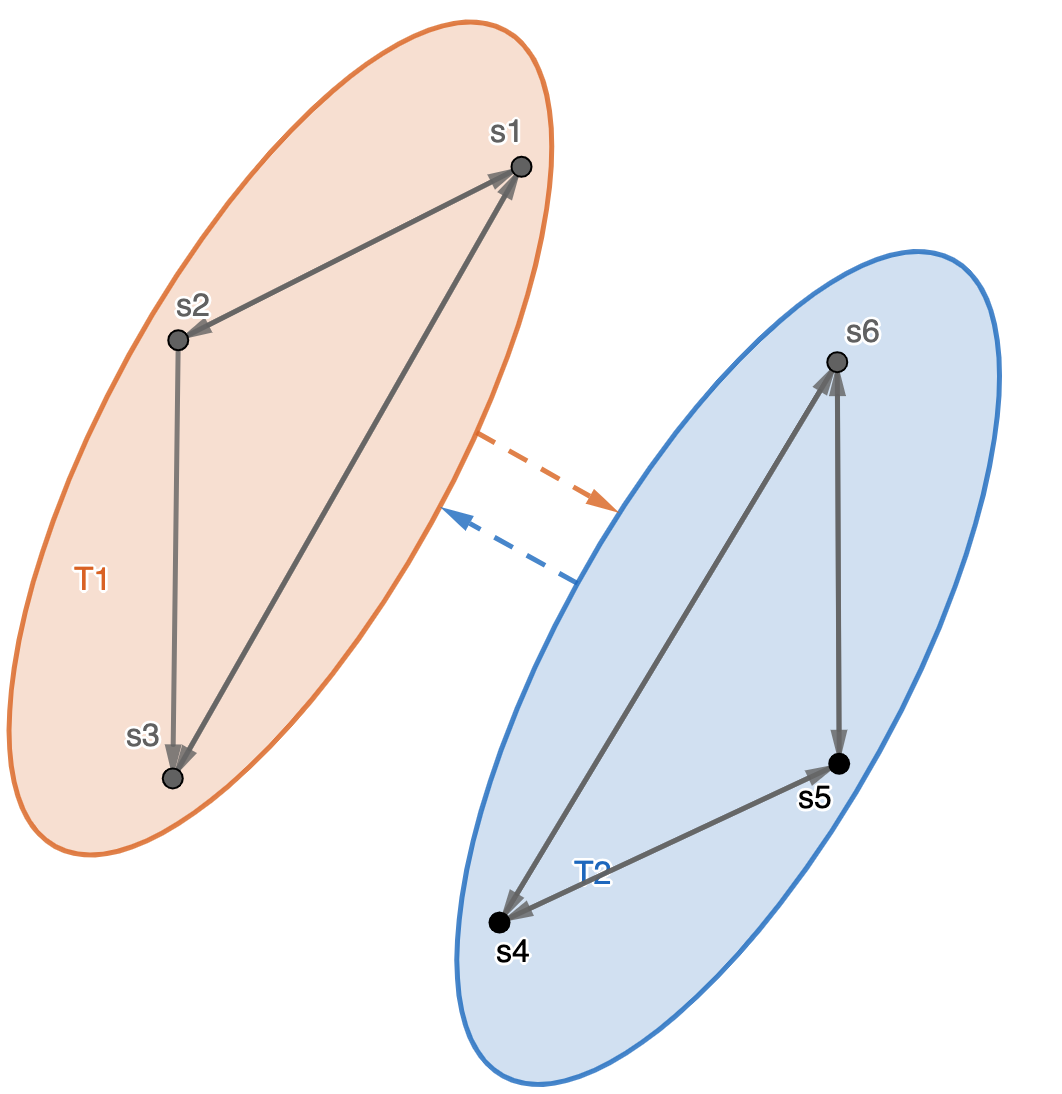} \label{sub-fig:hierarchy2}} \hfill
  \centering
  \subfigure[Section-sentence hierarchical addition (hierarchy-add, ours)]{\includegraphics[width=.35\textwidth]{plots/s3.png} \label{sub-fig:hierarchy3}} 
\caption{\small Comparison of the flat fully-connected graph used in \citet{erkan2004lexrank,mihalcea2004textrank,zheng2019sentence} to the 
hierarchical graph used in our models (b) and (c). Although the section-section multiplication reduces the edge computation proportionally to the number of sections, we found it oversimplifies the graph by assuming independence between sentences across different sections. Our final model loosens the assumption by including section-sentence connections as shown in sub-figure (c). }
\label{fig:three_hierarchy}
\end{figure}

\subsection{Different Hierarchical Structure}    
Besides our proposed hierarchical model (Figure \ref{fig:three_hierarchy} (c), hierarchy-add) in the paper, we also proposed and experimented with another novel hierarchical graph by introducing section-section  connections (Figure \ref{fig:three_hierarchy} (b), hierarchy-multiply). In this hierarchical setting, we multiply a sentence's sectional importance with its sentence importance (Eqn. (2)) to form the final centrality score:  

  \begin{equation}\label{eq:importance_mul}
        c(v^I_j) = \mu_1 \cdot c_{\text{inter}}(v^I_j) \times c_{\text{intra}}(v^I_j).
    \end{equation}

\begin{table}[ht!]
\centering
\resizebox{\columnwidth}{!}{%
\begin{tabular}{l c c c c}
\toprule
 \textbf{\scriptsize Model} & \textbf{\scriptsize ROUGE-1} &  \textbf{\scriptsize ROUGE-2} & \textbf{\scriptsize ROUGE-L} \\ 

\midrule
\multicolumn{4}{c}{Various Hierarchicall Centrality }\\
\midrule
no-hierarchy & 41.88& 15.39 &37.91\\
hierarchy-multiply (ours) &43.04 &16.76& 38.77 \\
hierarchy-add (ours) & \textbf{43.42} & \textbf{16.76} & \textbf{39.23}  \\
\bottomrule%
\end{tabular}
}
\caption{\label{tab:ablation2} \small Results on the PubMed validation set with different positional function or different hierarchical information.}
\end{table}

Our empirical results indicate the hierarchy-multiply model always outperforms no-hierarchy models ((Figure \ref{fig:three_hierarchy} (a)) but under performs hierarchy-add.
Nevertheless, Table \ref{tab:ablation2} shows that adding any hierarchical structure results in performance improvement by wide margins when compared to the no-hierarchy  model.

 \newpage  
 \begin{figure}[!hbt]
      \centering
      \subfigure[Oracle (arXiv)]{\includegraphics[width=.5\textwidth]{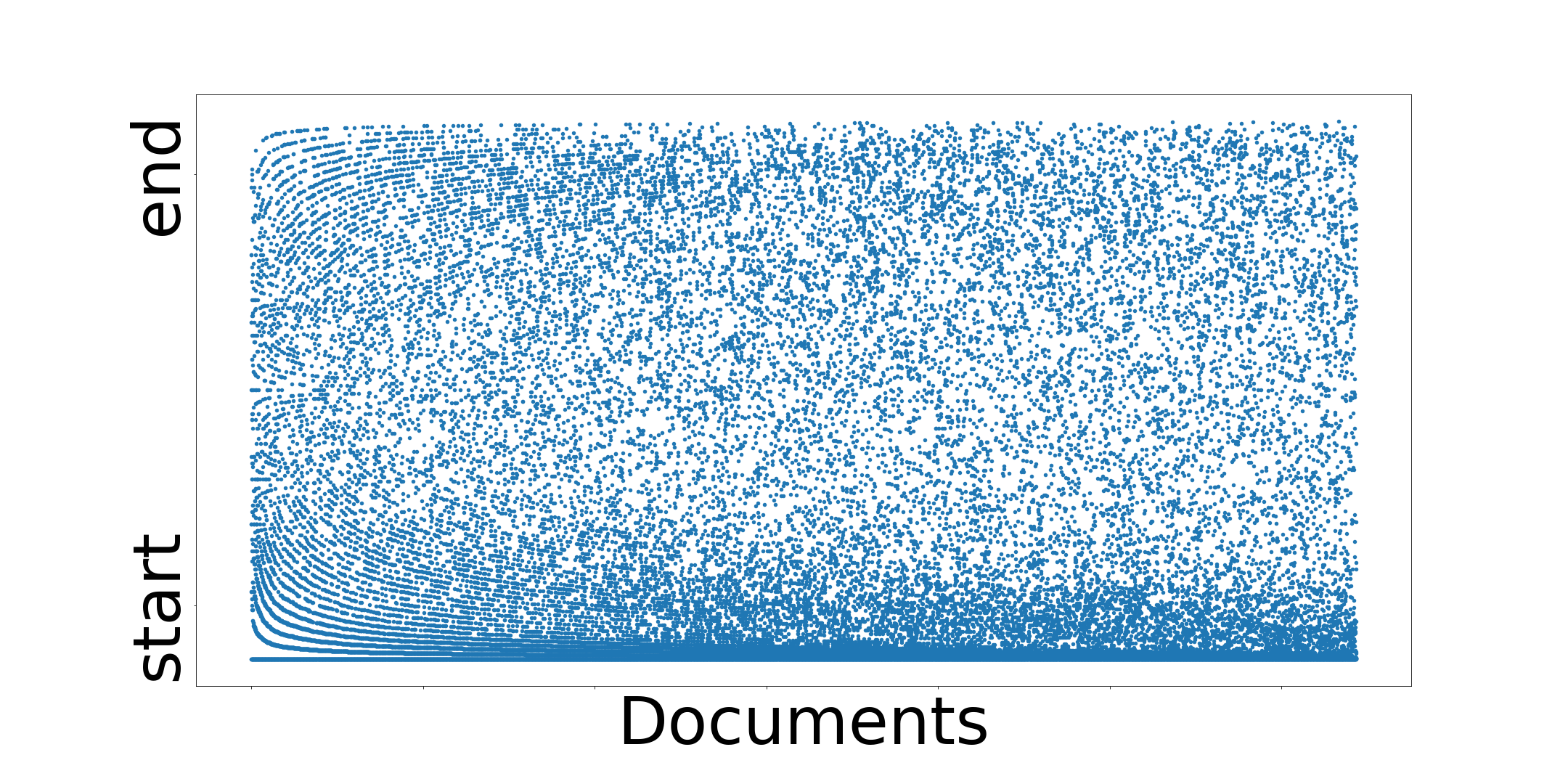} \label{sub-fig:arxiv1}} \hfill
      \subfigure[\textsc{PacSum} (arXiv)]{\includegraphics[width=.5\textwidth]{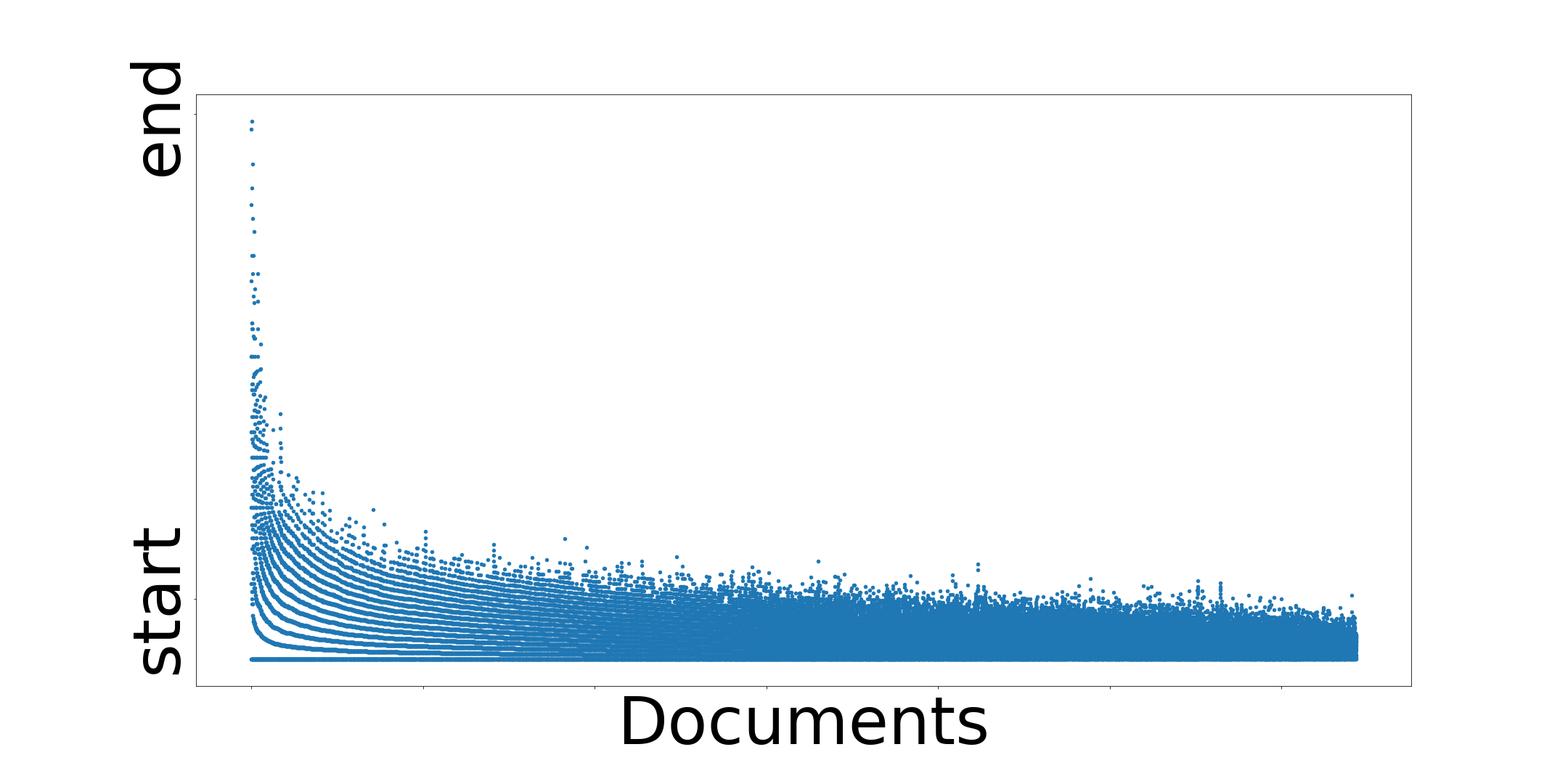} \label{sub-fig:arxiv2}} \hfill
      \subfigure[\textsc{HipoRank} (arXiv)]{\includegraphics[width=.5\textwidth]{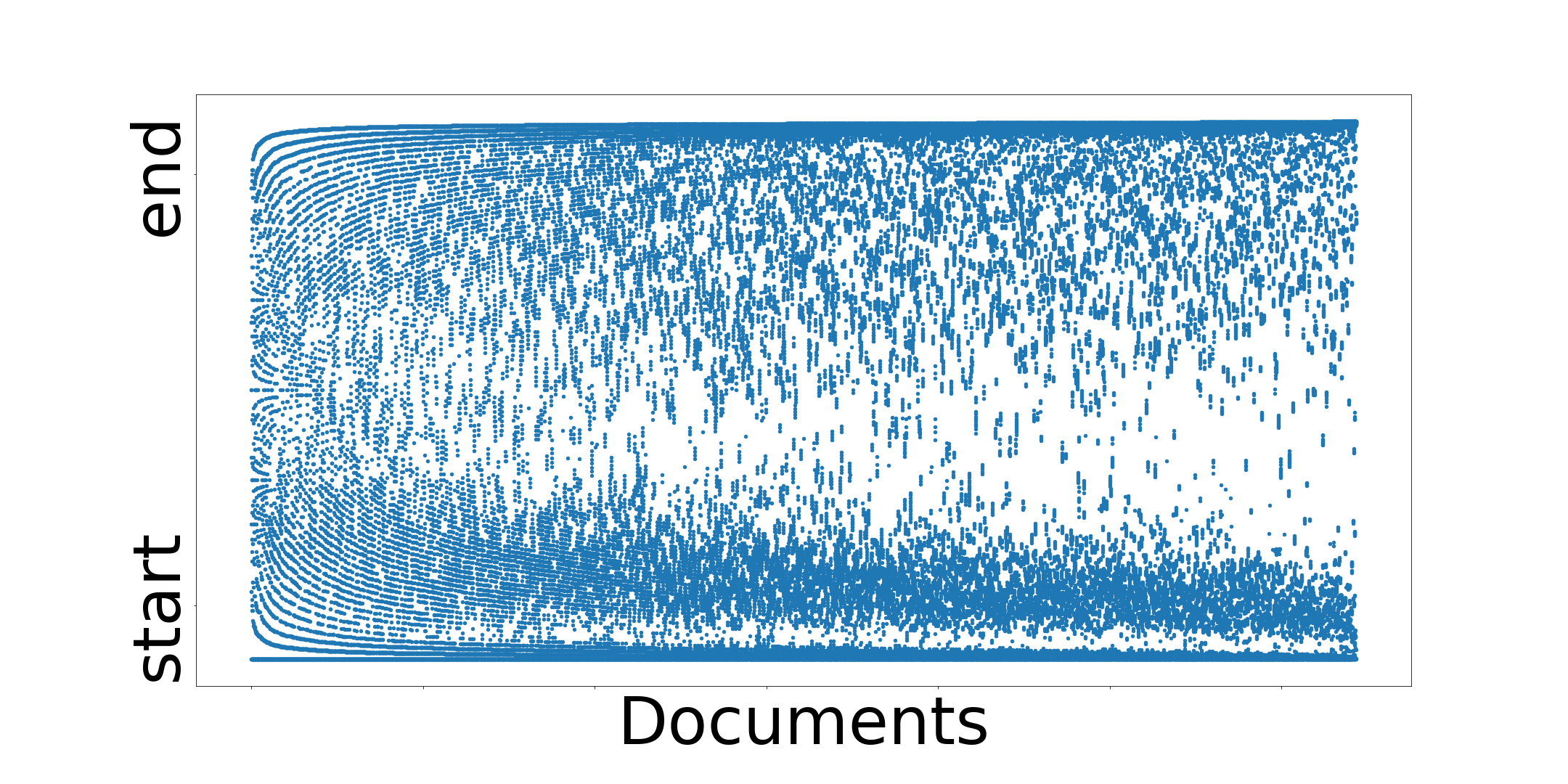} \label{sub-fig:arxiv3}} 
      \hfill
    \caption{Sentence positions in source document for extractive summaries generated by different models on the arXiv validation set. Documents on the x-axis are ordered by increasing article length from shortest to longest.}
    \label{fig:oracle_density_plot_arxiv}
    \end{figure}
\subsection{Sentence Position Comparison}

Figure \ref{fig:oracle_density_plot_arxiv} shows the sentence positions in source document for extractive summaries generated by different models on the arXiv validation set. We can again see that \textsc{PacSum} is biased towards selecting sentences that appear at the beginning of a document while our  method selects sentences in every section and near the article boundaries, similar to the oracle.

\end{document}